\documentclass[11pt,a4paper]{article}
\usepackage[hyperref]{emnlp-ijcnlp-2019}
\usepackage{times}
\usepackage{latexsym}
\usepackage{listings}
\usepackage[colorinlistoftodos]{todonotes}

\usepackage{subcaption}
\usepackage{url}
\usepackage{multirow}
\usepackage{array}
\usepackage[utf8]{inputenc}
\newcommand{\rev}[2]{\textcolor{black}{#2}}
\aclfinalcopy 

\title{LINSPECTOR WEB: A Multilingual Probing Suite for Word Representations}

\author{Max Eichler, Gözde Gül Şahin \& Iryna Gurevych \\
Research Training Group AIPHES and UKP Lab\\
Department of Computer Science\\
Technische Universit\"at Darmstadt\\
Darmstadt, Germany \\
\texttt{max.eichler@gmail.com}, \texttt{\{sahin,gurevych\}@ukp.tu-darmstadt.de} 
}

\date{}

\begin{document}
\maketitle
\begin{abstract}
We present \textsc{Linspector Web}, an open source multilingual inspector to analyze word representations. Our system provides researchers working in low-resource settings with an easily accessible web based probing tool to gain quick insights into their word embeddings especially outside of the English language. To do this we employ 16 simple linguistic probing tasks such as gender, case marking, and tense for a diverse set of 28 languages. We support probing of static word embeddings along with pretrained AllenNLP models that are commonly used for NLP downstream tasks such as named entity recognition, natural language inference and dependency parsing. The results are visualized in a polar chart and also provided as a table. \textsc{Linspector Web} is available as an offline tool or at \url{https://linspector.ukp.informatik.tu-darmstadt.de}.
\end{abstract}

\section{Introduction}

Natural language processing (NLP) has seen great progress after the introduction of continuous, dense, low dimensional vectors to represent text. The field has witnessed the creation of various word embedding models such as mono-lingual~\citep{Mikolov:2013:DRW:2999792.2999959}, contextualized~\citep{peters:NAACL2018}, multi-sense~\citep{PilehvarCNC17} and dependency-based~\citep{LevyG14}; as well as adaptation and design of neural network architectures for a wide range of NLP tasks. Despite their impressive performance, interpreting, analyzing and evaluating such black-box models have been shown to be challenging, which even led to a set of workshop series~\cite{W18-5400}.  

Early works for evaluating word representations~\cite{Faruqui2014CommunityEA,Faruqui2014ImprovingVS,nayak2016evaluating} have mostly focused on English and used either the word similarity or a set of downstream tasks. However datasets for either of those tasks do not exist for many languages, word similarity tests do not necessarily correlate well with downstream tasks and evaluating embeddings on downstream tasks can be too computationally demanding for low-resource scenarios. To address some of these challenges, \citet{shi-etAl:ACL2016,adi2017fine,Veldhoen2016DiagnosticCR,senteval18} have introduced \emph{probing tasks}, a.k.a. \emph{diagnostic classifiers}, that take as input a representation generated by a fully trained neural model and output predictions for a linguistic feature of interest. Due to its simplicity and low computational cost, it has been employed by many studies summarized by \citet{BelinkovG19}, mostly focusing on English. Unlike most studies, \citet{Khn2015WhatsIA} introduced a set of multilingual probing tasks, however its scope has been limited to syntactic tests and 7 languages. More importantly it is not accessible as a web application and the source code does not have support to probe pretrained downstream NLP models out of the box.

	\begin{table*}[h!]
		\small
		\centering
	    \begin{tabular}{ |c|c|c|c|c|c|c|c|c| } 
	        \hline
	         & \#Lang & \#Task-Type & Web & Offline & Static & Models & Layers & Epochs  \\
	        \hline
	        \cite{Faruqui2014CommunityEA} & 4 & 10-WST & $\times$ & $\times$ & $\times$ & & & \\
	        \cite{nayak2016evaluating} & 1 & 7-DT & $\times$ & $\times$ & $\times$ & & & \\
	        \cite{Khn2015WhatsIA} & 7 & 7-PT & & $\times$ & $\times$ & & & \\
	        Ours & 28 & 16-PT & $\times$ & $\times$ & $\times$ & $\times$ & $\times$ & $\times$ \\
	        \hline
	    \end{tabular}
	    \caption{Features of previous evaluation applications compared to Ours (\textsc{Linspector Web}). \#Lang: Number of supported languages, \#Task-Type: Number and type of the tasks, where WST: Word similarity tasks, DT: Downstream Tasks, PT: Probing Tasks. Static: Static word embeddings and Models: Pretrained downstream models.}
	    \label{table:1}
	\end{table*}

Given the above information, most of the lower-resource non-English academic NLP communities still suffer from (1) the amount of required human and computational resources to search for the right model configuration, and (2) the lack of diagnostics tools to analyze their models to gain more insights into what is captured. Recently, \citet{csahin2019linspector} proposed 16 multilingual probing tasks along with the corresponding datasets and \rev{showed correlate well with certain downstream task performances.}{found correlations between certain probing and downstream tasks and demonstrated their efficacy as diagnostic tools.} 
In this paper, we employ these datasets to develop \textsc{Linspector Web} that is designed to help researchers with low-resources working on non-English languages to (1) analyze, interpret, and visualize various layers of their pretrained AllenNLP~\cite{Gardner2017AllenNLP} models at different epochs and (2) measure the performance of static word embeddings for language-specific linguistic properties. To the best of our knowledge, this is the first web application that (a) performs online probing; (b) enables users to upload their pretrained downstream task models to automatically analyze different layers and epochs; and (c) has support for \rev{more than 20 lan-guages.}{28 languages with some of them being extremely low-resource such as Quechuan.}

\section{Previous Systems}
\label{sec:prev}
A now retired evaluation suite for word embeddings was \url{wordvectors.org}~\cite{Faruqui2014CommunityEA}. The tool provided evaluation and visualization for antonyms, synonyms, and female-male similarity; and later it was updated to support German, French, and Spanish word embeddings~\cite{Faruqui2014ImprovingVS}. For a visualization the user could enter multiple tokens and would receive a 2 dimensional chart to visualize the cosine distance between the tokens. Therefore it was limited by the amount of tokens, a human could enter and analyze. VecEval~\cite{nayak2016evaluating} is another web based suite for static English word embeddings that perform evaluation on a set of downstream tasks which may take several hours. The visualization is similar to \textsc{Linspector Web} reporting both charts and a table. Both web applications do not support probing of intermediate layers of pretrained models or the addition of multiple epochs. 
\citet{Khn2015WhatsIA} introduced an offline, multilingual probing suite for static embeddings limited in terms of the languages and the probing tasks. A comparison of the system features of previous studies is given in Table~\ref{table:1}.

\section{\textsc{Linspector Web}}
\label{sec:model}
	\begin{figure*}
	  \centering
	    \begin{subfigure}[b]{0.32\textwidth}
	      \includegraphics[width=\textwidth]{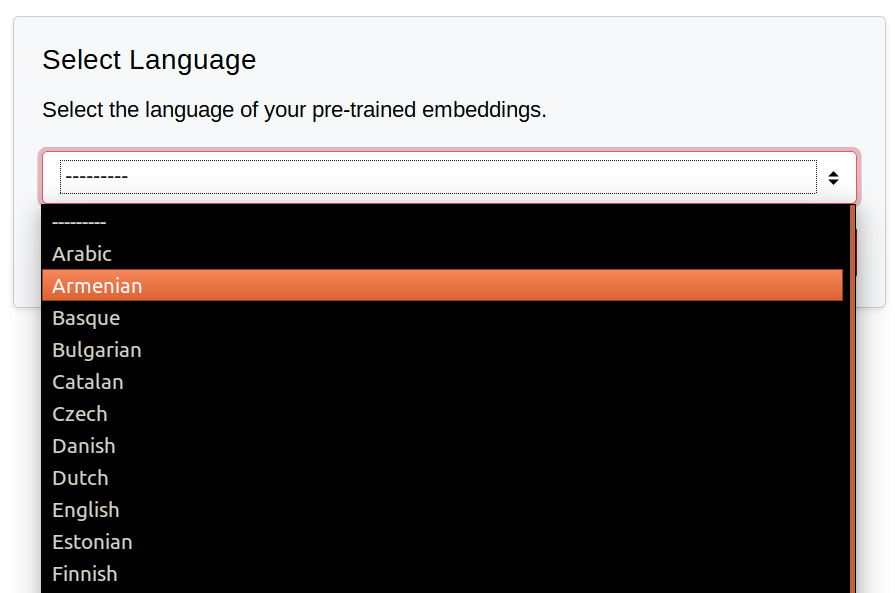}
	  \end{subfigure}
      ~
	  \begin{subfigure}[b]{0.31\textwidth}
	      \includegraphics[width=\textwidth]{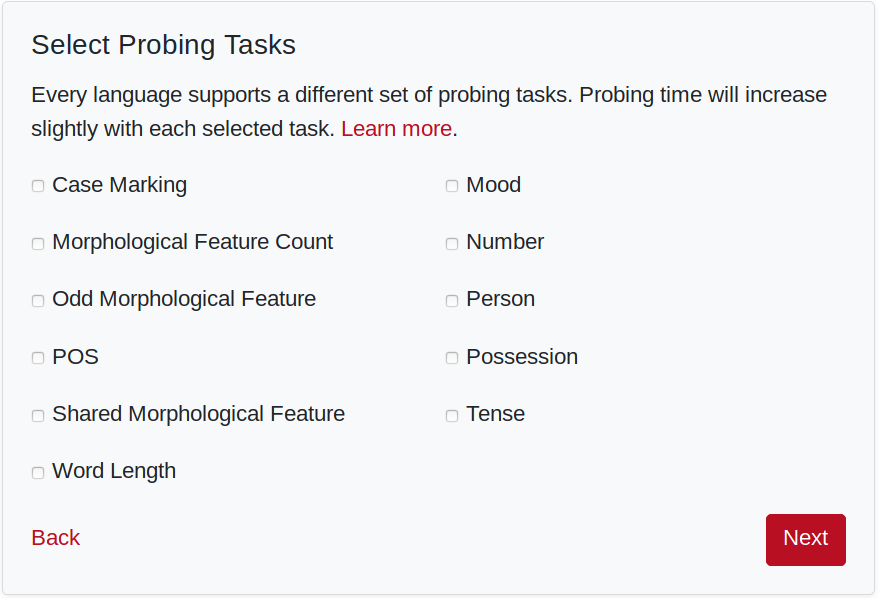}
	  \end{subfigure}
	  ~
	  \begin{subfigure}[b]{0.32\textwidth}
	      \includegraphics[width=\textwidth]{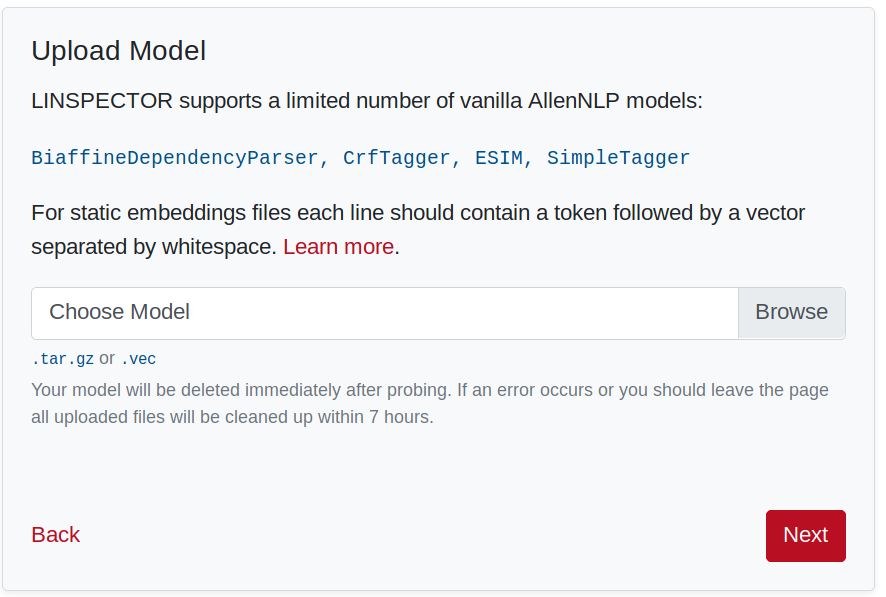}
	  \end{subfigure}
	  \caption{\textbf{Left}: Language selection, \textbf{Middle}: Probing task selection,  \textbf{Right}: Uploading model.}
	\label{fig:pipeline}
	\end{figure*}
  	Our system is targeted at multilingual researchers working with low-resource settings. 
  	It is designed as a web application to enable such users to probe their word representations with minimal effort and computational requirements by simply uploading a file. The users can either upload their \textbf{pretrained static embeddings files} (e.g. word2vec~\cite{Mikolov:2013:DRW:2999792.2999959}, fastText~\cite{DBLP:journals/corr/BojanowskiGJM16}, GloVe~\cite{pennington-etal-2014-glove});~\footnote{\tiny Since our probing datasets are publicly available, fastText embeddings for unknown words in our dataset can be generated by the user locally via the provided functions in \cite{csahin2019linspector}.}or their \textbf{pretrained archived AllenNLP models}. 
  	In this version, we only give support to AllenNLP, due to its high usage rate by low-resource community and being up-to-date, i.e., containing state-of-the-art models for many NLP tasks and being continuously maintained at the Allen Institute for Artificial Intelligence~\cite{Gardner2017AllenNLP}. 
  	
  	\subsection{Scope of Probing}
	\label{ssec:scope}
		We support 28 languages from very diverse language families.~\footnote{\tiny Arabic, Armenian, Basque, Bulgarian, Catalan, Czech, Danish, Dutch, English, Estonian, Finnish, French, German, Greek, Hungarian, Italian, Macedonian, Polish, Portuguese, Quechuan, Romanian, Russian, Serbian, Serbo-Crotian, Spanish, Swedish, Turkish and Vietnamese} The multilingual probing datasets~\cite{csahin2019linspector} used in this system are language-specific, i.e., languages with a gender system are probed for gender, whereas languages with a rich case-marking system are probed for case. The majority of the probing tasks probe for morpho-syntactic properties (e.g. case, mood, person) which have been shown to correlate well with syntactic and semantic parsing for a number of languages, where a small number of tasks probe for surface (e.g. word length) or semantic level properties (e.g. pseudoword). Finally, there are two morphological comparison tasks (Odd/Shared Morphological Feature) aiming to find the unique distinct/shared morphological feature between two tokens, which have been shown to correlate well with the NLI task. The current probing tasks are 
		type-level (i.e. do not contain ambiguous words) and are filtered to keep only the frequent words. These tasks are (1) domain independent and (2) contain valuable information encoded via subwords in many languages (e.g. the Turkish word \emph{gelemeyenlerden} ``he/she is one of the folks who can not come'' encodes sentence-level information). \rev{}{\footnote{Users can choose probing tasks either intuitively or rely on earlier studies e.g., that show a linguistic feature has been beneficial for the downstream task. Therefore not every probing task is needed during a specific evaluation.}}
	
	\subsection{Features: Models, Layers and Epochs}
	\label{ssec:feat}
		\begin{figure*}
		  \centering
		    \begin{subfigure}[b]{0.45\textwidth}
		      \includegraphics[width=\textwidth]{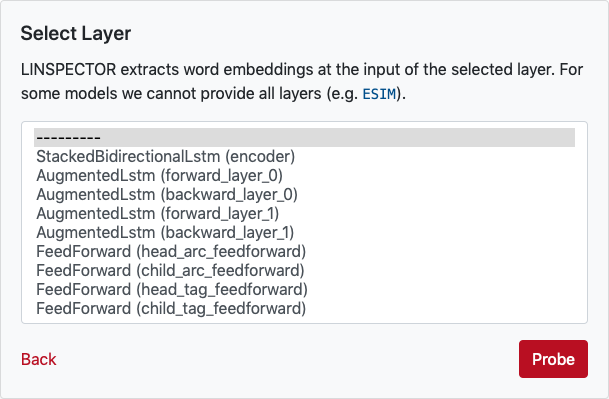}
		  \end{subfigure}
	      ~
		  \begin{subfigure}[b]{0.45\textwidth}
		      \includegraphics[width=\textwidth]{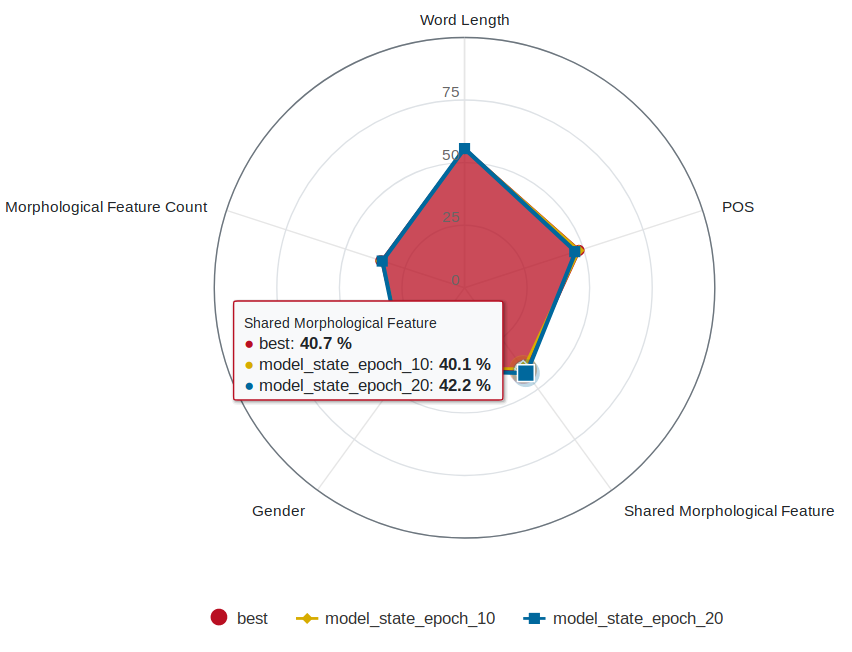}
		  \end{subfigure}
		  \caption{\textbf{Left}: Layer selection example, \textbf{Right}: Polar chart result shown for different epochs for pretrained Arabic BiaffineDependencyParser.}
		\label{fig:feats}
		\end{figure*}
		We support the following classifier architectures implemented by AllenNLP: BiaffineDependencyParser~\cite{DBLP:journals/corr/DozatM16}, CrfTagger~\cite{Lafferty2001ConditionalRF}, SimpleTagger~\cite{Gardner2017AllenNLP}, ESIM~\cite{Chen2017EnhancedLF}. BiaffineDependencyParser and CrfTagger are highlighted as the default choice for dependency parsing and named entity recognition by \cite{Gardner2017AllenNLP}, while ESIM was picked as one of two available natural language inference models, and SimpleTagger support was added as the entry level AllenNLP classifier to solve tasks like parts-of-speech tagging. 

		The users can choose the layers they want to probe. This allows the users to analyze what linguistic information is captured by different layers of the model \textit{(e.g., POS information in lower layers, semantic information in higher levels)}. It is possible to select any AllenNLP encoder layer for classifiers with token, sentence, or document based input and models with dual input (e.g. ESIM: premise, hypothesis) that allow probing of selected layers depending on their internal architecture as described in Sec.~\ref{ssec:backend}. Additionally a user can specify up to 3 epochs for probing to inspect what their model learns and forgets during training. This is considered a crucial feature \rev{}{as it provides insights on learning dynamics of models~\cite{DBLP:conf/naacl/SaphraL19}}. For instance, a user diagnosing a pretrained NLI task, can probe for the tasks that have been shown to correlate well (Mood, Person, Polarity, and Tense)~\cite{csahin2019linspector} for additional epochs, and analyze how their performance evolves during training. After the diagnostic classifiers are trained and tested on the specified language, model, layer, and epochs, the users are provided with (1) accuracies of each task visualized in a polar chart, (2) a table containing accuracy and loss for each probing test, and (3) in case of additional epochs, accuracies for other epochs are overlaid on the chart and columns are added to the table for easy comparison as shown in Fig.~\ref{fig:feats}-Right. 

		The uploaded model files are deleted immediately after probing, however the results can be saved or shared via a publicly accessible URL. The project is open source and easily extendable to additional languages, probing tasks and AllenNLP models. New languages can be added simply by adding train, dev, and test data for selected probing tasks and adding one database entry. Similarly new probing tasks can be defined following \cite{csahin2019linspector}. In case the new tasks differ by input type, a custom AllenNLP dataset reader and classifier should be added. It can be extended to new AllenNLP models by adding the matching predictor to the supported list or writing a custom predictor if the model requires dual input values (e.g. ESIM). Finally, other frameworks (e.g. ONNX format) can be supported by adding a method to extract embeddings from the model.

\section{System Description}
	\textsc{Linspector Web} is based on the Python Django framework~\footnote{\tiny \url{https://www.djangoproject.com}} which manages everything related to performance, security, scalability, and database handling.

	\subsection{Frontend}
	\label{ssec:frontend}
		First, the user selects the language of the model and a number of probing tests they want to perform. The probing test selection menu will vary with the selected language. Next the user has to upload an archived AllenNLP model or a static embeddings file. The input pipeline is shown in Fig.~\ref{fig:pipeline}. The upload is handled asynchronously using custom AJAX code to support large files, prevent timeouts, and give the user some progress feedback. 
		The backend detects if an uploaded file is an archived AllenNLP model and provides a list of layers if that is the case as shown in Fig.~\ref{fig:feats}-Left. Probing is handled asynchronously by the backend. A JSON API endpoint gives progress feedback to the frontend which displays a progress bar and the currently executed probing test to the user. Finally results are displayed in an interactive chart and a table. For the user interface, we use the Bootstrap framework~\footnote{\tiny \url{https://getbootstrap.com}} that provides us with modern, responsive, and mobile compatible HTML and CSS. The visualization is done using the Highcharts library.~\footnote{\tiny \url{https://www.highcharts.com}}

	\subsection{Backend}
	\label{ssec:backend}
\begin{figure}
		  \centering
		  \includegraphics[width={0.25\textwidth}]{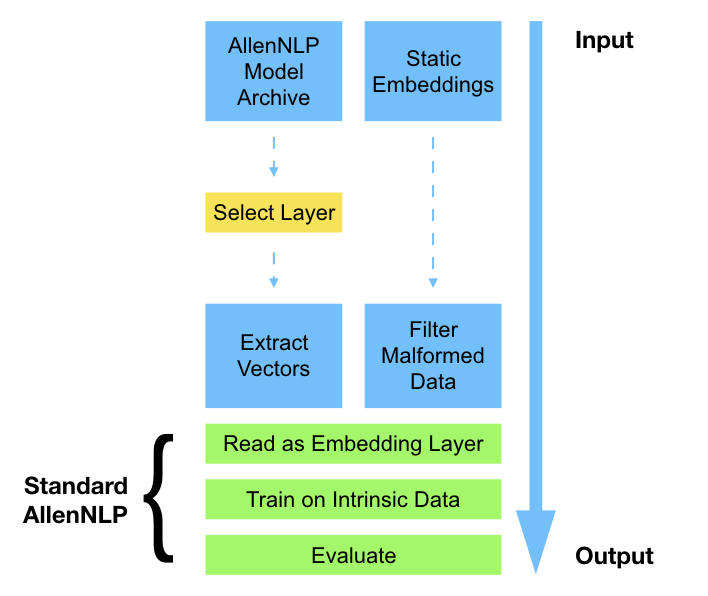} 
		  \caption{Backend architecture}
		\label{fig:sys_arch}
		\end{figure}

		The structure of the backend system is shown in Fig.~\ref{fig:sys_arch} and the main components are explained below.
		\paragraph{Layers:}To get a list of layers an archived AllenNLP model is loaded using a standard AllenNLP API. Since every AllenNLP classifier inherits from the PyTorch~\cite{paszke2017automatic} class \texttt{torch.nn.Module}, we \rev{can get a list of}{gather two levels of immediate} submodules using the \texttt{named\_children()} API. First we extract high level AllenNLP modules including all \texttt{Seq2SeqEncoder}, \texttt{Seq2VecEncoder}, and \texttt{FeedForward} modules by testing each submodule for a \texttt{get\_input\_dim()} method. Then we extract low level modules which can be either AllenNLP modules e.g. \texttt{AugmentedLstm} or PyTorch modules e.g. \texttt{Linear} by testing for the attributes \texttt{input\_size} or \texttt{in\_features}. All those modules are then returned as available probing layers. We require the input dimension later and since there is no standard API we have to exclude some submodules. \rev{}{\footnote{By analyzing the AllenNLP codebase and PyTorch modules used in their sample configurations, we decided to support modules providing their dimension using \texttt{get\_input\_dim()}, \texttt{input\_size}, or \texttt{in\_features}; which may change with future versions of AllenNLP.}} Finally we don't support models that require additional linguistic information such as POS tags. 
		\paragraph{Getting Embeddings:} PyTorch modules allow us to register forward hooks. A hook is a callback which receives the module, input, and output every time an input is passed through the module. For AllenNLP models we register such a callback to the selected encoding layer. Then each time a token is passed through the model, it passes through the encoder and the callback receives the input vector. The most reliable way to pass tokens through a model is using AllenNLP predictors. There is a matching predictor for every model which are regularly tested and updated. We gather all tokens from our intrinsic probing data and pass it through the predictor. For every token the forward hook is called in the background which then provides us with the vector. The token and vector are then written to a temporary file. During the embedding extraction, the progress is reported back to the frontend periodically in 30 steps. For static embeddings all lines that match the embedding dimension are written to a temporary file and malformed data is removed. 
		\paragraph{Probing:} Finally the gathered embeddings are loaded as a pretrained non-trainable embedding layer for a single linear layer custom build AllenNLP classifier. The classifier is trained and evaluated using the intrinsic probing data for the specified probing test. We use 20 epochs with early stopping, a patience of 5, and gradient clipping of 0.5. The evaluation accuracy and loss are then returned. For contrastive probing tasks (Odd/Shared Morphological Feature) a similar linear classifier that takes concatenated tokens as input, is used. 
		\paragraph{Asynchronous} probing is handled using the Python Celery framework~\footnote{\tiny\url{http://www.celeryproject.org}}, the RabbitMQ message broker~\footnote{\tiny\url{https://www.rabbitmq.com}}, and the Eventlet execution pool~\footnote{\tiny\url{https://eventlet.net}}. When the user starts probing, a new Celery task is created in the backend which executes all probing tasks specified by the user asynchronously and reports the progress back to the frontend. Finally the results are saved in a PostgreSQL 
		or SQLite
		 database using the Django Celery Results application.

	\subsection{Tests}
		We have trained BiaffineDependencyParser, CrfTagger, SimpleTagger, and ESIM AllenNLP models for Arabic, Armenian, Czech, French, and Hungarian with varying dimensions. We have tested the intrinsic probing data, layer selection, consistency of metrics, contrastive and non-contrastive classifiers, and all probing tests for multiple combinations of languages, dimensions, and AllenNLP models. Static embeddings are tested using pretrained fastText files for the same languages. 
		In addition, the file upload was tested with files up to 8 GB over a DSL 50k connection.

	\subsection{Training Times} The \textsc{Linspector Web} server is hosted in university data center with a state-of-the-art internet connection which allows for fast upload speeds. Therefore, the overall upload speed mostly depends on the users connection. For a single probing task, embedding extraction, training, and evaluation is around a few minutes.


\section{Conclusion}
\label{sec:conc}
	Researchers working on non-English languages under low-resource settings have lacked a tool that would assist with model selection via providing linguistic insights, to this date. To address this, we presented \textsc{Linspector Web}, an open source, web-based evaluation suite with 16 probing tasks for 28 languages; which can probe pretrained static word embeddings and various layers of a number of selected AllenNLP models. The tool can easily be extended for additional languages, probing tasks, probing models and AllenNLP models. \textsc{Linspector Web} is available at \url{https://linspector.ukp.informatik.tu-darmstadt.de} and the source code for the server is released with \url{https://github.com/UKPLab/linspector-web} along with the installation instructions for the server. \rev{Probing tasks and the system will be extended to support contextual probing in near future.}{The system is currently being extended to support contextualized word embeddings with contextualized probing tasks using the Universal Dependency Treebanks~\cite{11234/1-2988}.}

\section{Acknowledgements}
\label{sec:ack}
This work has been supported by the DFG-funded research training group “Adaptive Preparation of Information form Heterogeneous Sources” (AIPHES, GRK 1994/1), and also by the German Federal Ministry of Education and Research (BMBF) under the promotional reference 01UG1816B (CEDIFOR) and as part of the Software Campus program under the promotional reference 01IS17050. 

\bibliography{emnlp-ijcnlp-2019}

\begin{thebibliography}{24}
\expandafter\ifx\csname natexlab\endcsname\relax\def\natexlab#1{#1}\fi

\bibitem[{Adi et~al.(2017)Adi, Kermany, Belinkov, Lavi, and
  Goldberg}]{adi2017fine}
Yossi Adi, Einat Kermany, Yonatan Belinkov, Ofer Lavi, and Yoav Goldberg. 2017.
\newblock Fine-grained analysis of sentence embeddings using auxiliary
  prediction tasks.
\newblock In \emph{{ICLR} 2017, Toulon, France, April 24-26, 2017, Conference
  Track Proceedings}.

\bibitem[{Belinkov and Glass(2019)}]{BelinkovG19}
Yonatan Belinkov and James Glass. 2019.
\newblock Analysis methods in neural language processing: {A} survey.
\newblock \emph{{TACL}}, 7:49--72.

\bibitem[{Bojanowski et~al.(2016)Bojanowski, Grave, Joulin, and
  Mikolov}]{DBLP:journals/corr/BojanowskiGJM16}
Piotr Bojanowski, Edouard Grave, Armand Joulin, and Tomas Mikolov. 2016.
\newblock \href {http://arxiv.org/abs/1607.04606} {Enriching word vectors with
  subword information}.
\newblock \emph{CoRR}, abs/1607.04606.

\bibitem[{Chen et~al.(2017)Chen, Zhu, Ling, Wei, Jiang, and
  Inkpen}]{Chen2017EnhancedLF}
Qian Chen, Xiaodan Zhu, Zhen{-}Hua Ling, Si~Wei, Hui Jiang, and Diana Inkpen.
  2017.
\newblock Enhanced {LSTM} for natural language inference.
\newblock In \emph{{ACL} 2017, Vancouver, Canada, July 30 - August 4, Volume 1:
  Long Papers}, pages 1657--1668.

\bibitem[{Conneau et~al.(2018)Conneau, Kruszewski, Lample, Barrault, and
  Baroni}]{senteval18}
Alexis Conneau, Germ{\'{a}}n Kruszewski, Guillaume Lample, Lo{\"{\i}}c
  Barrault, and Marco Baroni. 2018.
\newblock What you can cram into a single
  {\textbackslash}{\textdollar}{\&}!{\#}* vector: Probing sentence embeddings
  for linguistic properties.
\newblock In \emph{{ACL} 2018, Melbourne, Australia, July 15-20, 2018, Volume
  1: Long Papers}, pages 2126--2136.

\bibitem[{Dozat and Manning(2016)}]{DBLP:journals/corr/DozatM16}
Timothy Dozat and Christopher~D. Manning. 2016.
\newblock \href {http://arxiv.org/abs/1611.01734} {Deep biaffine attention for
  neural dependency parsing}.
\newblock \emph{CoRR}, abs/1611.01734.

\bibitem[{Faruqui and Dyer(2014{\natexlab{a}})}]{Faruqui2014CommunityEA}
Manaal Faruqui and Chris Dyer. 2014{\natexlab{a}}.
\newblock Community evaluation and exchange of word vectors at wordvectors.org.
\newblock In \emph{{ACL} 2014, June 22-27, 2014, Baltimore, MD, USA, System
  Demonstrations}, pages 19--24.

\bibitem[{Faruqui and Dyer(2014{\natexlab{b}})}]{Faruqui2014ImprovingVS}
Manaal Faruqui and Chris Dyer. 2014{\natexlab{b}}.
\newblock Improving vector space word representations using multilingual
  correlation.
\newblock In \emph{{EACL} 2014, April 26-30, 2014, Gothenburg, Sweden}, pages
  462--471.

\bibitem[{Gardner et~al.(2018)Gardner, Grus, Neumann, Tafjord, Dasigi, Liu,
  Peters, Schmitz, and Zettlemoyer}]{Gardner2017AllenNLP}
Matt Gardner, Joel Grus, Mark Neumann, Oyvind Tafjord, Pradeep Dasigi, Nelson
  Liu, Matthew Peters, Michael Schmitz, and Luke Zettlemoyer. 2018.
\newblock \href {http://arxiv.org/abs/1803.07640} {Allennlp: A deep semantic
  natural language processing platform}.

\bibitem[{K{\"{o}}hn(2015)}]{Khn2015WhatsIA}
Arne K{\"{o}}hn. 2015.
\newblock What's in an embedding? analyzing word embeddings through
  multilingual evaluation.
\newblock In \emph{{EMNLP} 2015, Lisbon, Portugal, September 17-21, 2015},
  pages 2067--2073.

\bibitem[{Levy and Goldberg(2014)}]{LevyG14}
Omer Levy and Yoav Goldberg. 2014.
\newblock Dependency-based word embeddings.
\newblock In \emph{{ACL} 2014, June 22-27, 2014, Baltimore, MD, USA, Volume 2:
  Short Papers}, pages 302--308.

\bibitem[{Linzen et~al.(2018)Linzen, Chrupa{\l}a, and Alishahi}]{W18-5400}
Tal Linzen, Grzegorz Chrupa{\l}a, and Afra Alishahi. 2018.
\newblock Proceedings of the 2018 emnlp workshop blackboxnlp: Analyzing and
  interpreting neural networks for nlp.
\newblock Association for Computational Linguistics.

\bibitem[{Mikolov et~al.(2013)Mikolov, Sutskever, Chen, Corrado, and
  Dean}]{Mikolov:2013:DRW:2999792.2999959}
Tomas Mikolov, Ilya Sutskever, Kai Chen, Greg Corrado, and Jeffrey Dean. 2013.
\newblock Distributed representations of words and phrases and their
  compositionality.
\newblock In \emph{{NIPS} 2013 - Volume 2}, NIPS'13, pages 3111--3119.

\bibitem[{Nayak et~al.(2016)Nayak, Angeli, and Manning}]{nayak2016evaluating}
Neha Nayak, Gabor Angeli, and Christopher~D Manning. 2016.
\newblock Evaluating word embeddings using a representative suite of practical
  tasks.
\newblock In \emph{{RepEval}{@ACL}}, pages 19--23.

\bibitem[{Nivre et~al.(2019)Nivre, Abrams, Željko Agić, and
  et~al.}]{11234/1-2988}
Joakim Nivre, Mitchell Abrams, Željko Agić, and et~al. 2019.
\newblock Universal dependencies 2.4.
\newblock {LINDAT}/{CLARIN} digital library at the Institute of Formal and
  Applied Linguistics ({{\'U}FAL}), Faculty of Mathematics and Physics, Charles
  University.

\bibitem[{Paszke et~al.(2017)Paszke, Gross, Chintala, Chanan, Yang, DeVito,
  Lin, Desmaison, Antiga, and Lerer}]{paszke2017automatic}
Adam Paszke, Sam Gross, Soumith Chintala, Gregory Chanan, Edward Yang, Zachary
  DeVito, Zeming Lin, Alban Desmaison, Luca Antiga, and Adam Lerer. 2017.
\newblock \href {https://openreview.net/forum?id=BJJsrmfCZ} {Automatic
  differentiation in {PyTorch}}.
\newblock In \emph{NIPS Autodiff Workshop}.

\bibitem[{Pennington et~al.(2014)Pennington, Socher, and
  Manning}]{pennington-etal-2014-glove}
Jeffrey Pennington, Richard Socher, and Christopher Manning. 2014.
\newblock {G}love: Global vectors for word representation.
\newblock In \emph{{EMNLP})}, pages 1532--1543, Doha, Qatar. Association for
  Computational Linguistics.

\bibitem[{Peters et~al.(2018)Peters, Neumann, Iyyer, Gardner, Clark, Lee, and
  Zettlemoyer}]{peters:NAACL2018}
Matthew~E. Peters, Mark Neumann, Mohit Iyyer, Matt Gardner, Christopher Clark,
  Kenton Lee, and Luke Zettlemoyer. 2018.
\newblock Deep contextualized word representations.
\newblock In \emph{{NAACL-HLT} 2018, New Orleans, Louisiana, USA, June 1-6,
  2018, Volume 1 (Long Papers)}, pages 2227--2237.

\bibitem[{Pilehvar et~al.(2017)Pilehvar, Camacho{-}Collados, Navigli, and
  Collier}]{PilehvarCNC17}
Mohammad~Taher Pilehvar, Jos{\'{e}} Camacho{-}Collados, Roberto Navigli, and
  Nigel Collier. 2017.
\newblock Towards a seamless integration of word senses into downstream {NLP}
  applications.
\newblock In \emph{{ACL} 2017, Vancouver, Canada, July 30 - August 4, Volume 1:
  Long Papers}, pages 1857--1869.

\bibitem[{{\c{S}}ahin et~al.(2019){\c{S}}ahin, Vania, Kuznetsov, and
  Gurevych}]{csahin2019linspector}
G{\"o}zde~G{\"u}l {\c{S}}ahin, Clara Vania, Ilia Kuznetsov, and Iryna Gurevych.
  2019.
\newblock Linspector: Multilingual probing tasks for word representations.
\newblock \emph{arXiv preprint arXiv:1903.09442}.

\bibitem[{Saphra and Lopez(2019)}]{DBLP:conf/naacl/SaphraL19}
Naomi Saphra and Adam Lopez. 2019.
\newblock \href {https://aclweb.org/anthology/papers/N/N19/N19-1329/}
  {Understanding learning dynamics of language models with {SVCCA}}.
\newblock In \emph{{NAACL-HLT} 2019, Minneapolis, MN, USA, June 2-7, 2019,
  Volume 1 (Long and Short Papers)}, pages 3257--3267.

\bibitem[{Shi et~al.(2016)Shi, Padhi, and Knight}]{shi-etAl:ACL2016}
Xing Shi, Inkit Padhi, and Kevin Knight. 2016.
\newblock {Does String-Based Neural {MT} Learn Source Syntax?}
\newblock In \emph{{EMNLP} 2016, Austin, Texas, USA, November 1-4, 2016}, pages
  1526--1534.

\bibitem[{Sutton et~al.(2007)Sutton, McCallum, and
  Rohanimanesh}]{Lafferty2001ConditionalRF}
Charles~A. Sutton, Andrew McCallum, and Khashayar Rohanimanesh. 2007.
\newblock Dynamic conditional random fields: Factorized probabilistic models
  for labeling and segmenting sequence data.
\newblock \emph{Journal of Machine Learning Research}, 8:693--723.

\bibitem[{Veldhoen et~al.(2016)Veldhoen, Hupkes, and
  Zuidema}]{Veldhoen2016DiagnosticCR}
Sara Veldhoen, Dieuwke Hupkes, and Willem~H. Zuidema. 2016.
\newblock Diagnostic classifiers revealing how neural networks process
  hierarchical structure.
\newblock In \emph{Proceedings of the Workshop on Cognitive Computation:
  Integrating neural and symbolic approaches {@}{NIPS} 2016), Barcelona, Spain,
  December 9, 2016.}

\end{thebibliography}
\bibliographystyle{acl_natbib}

\end{document}